# *PhyPlan*: Compositional and Adaptive Physical Task Reasoning with Physics-Informed Skill Networks for Robot Manipulators

Harshil Vagadia[2,*], Mudit Chopra[1,*], Abhinav Barnawal[1,+], Tamajit Banerjee[2,+], Shreshth Tuli[2], Souvik Chakraborty[1], and Rohan Paul[1]

[1]Work primarily done when at IIT Delhi, [2]Affiliated with IIT Delhi, [*,+]Indicate equal contributions

**Given the task of positioning a ball-like object to a goal region beyond direct reach, humans can often throw, slide, or rebound objects against the wall to attain the goal. However, enabling robots to reason similarly is non-trivial. Existing methods for physical reasoning are data-hungry and struggle with complexity and uncertainty inherent in the real world. This paper presents PhyPlan, a novel physics-informed planning framework that combines physics-informed neural networks (PINNs) with modified Monte Carlo Tree Search (MCTS) to enable embodied agents to perform dynamic physical tasks. PhyPlan leverages PINNs to simulate and predict outcomes of actions in a fast and accurate manner and uses MCTS for planning. It dynamically determines whether to consult a PINN-based simulator (coarse but fast) or engage directly with the actual environment (fine but slow) to determine optimal policy. Evaluation with robots in simulated 3D environments demonstrates the ability of our approach to solve 3D-physical reasoning tasks involving the composition of dynamic skills. Quantitatively, PhyPlan excels in several aspects: (i) it achieves lower regret when learning novel tasks compared to state-of-the-art, (ii) it expedites skill learning and enhances the speed of physical reasoning, (iii) it demonstrates higher data efficiency compared to a physics un-informed approach.**

Physics-informed networks | Model-based RL | Robot Manipulation

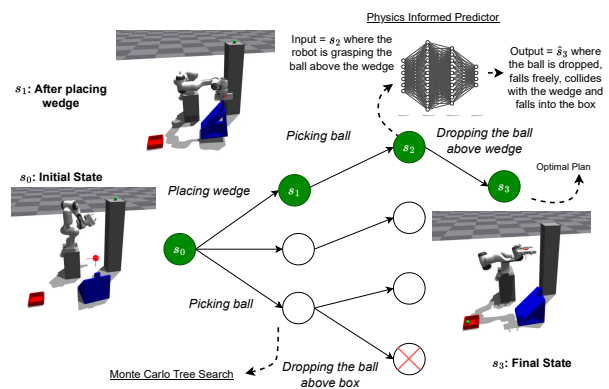

**Fig. 1. Approach Overview.** Given the task of putting the green ball into the goal region (red box), PhyPlan decomposes the planning problem into generating physics-informed priors (using PINN) to facilitate action search using MCTS to reach the intended goal region. We find the optimal plan where we bounce a ball off a wedge instead of directly dropping it on the box, which is unreachable.

## Introduction

Consider the scenario where the robot is asked to put a ball inside an empty box. The robot must improvise and dynamically interact with the objects in the environment if the box is far away. It may need to throw the ball, slide it across, or bounce it off a wedge to make it reach the box. Reasoning about the effects of various actions is a challenging task. Although the solution may appear self-evident to humans, a robot requires several demonstrations of each alternative to attain proficiency in a given task. Learning to reason with physical skills is a hallmark of intelligence. Increasing evidence is emerging that humans possess an intuitive physics engine involved in perceptual and goal-directed reasoning (1, 2). In essence, such reasoning requires (implicit or explicit) learning of predictive models of skills and the ability to chain them to attain a stated goal.

Recent efforts such as (1) learn to solve physical tasks by exploring interactions in a physics simulator. The reliance on a realistic physical simulator during training slows the learning process. Others, such as (3) benchmark, propose alternative model-free approaches and show their limitation in long-horizon reasoning, mainly when rewards are sparse (using a physical tool in a specific way to acquire large rewards). This motivates an investigation into the availability and use of "physical priors" to scale skill learning to longer horizons. The work of (4) addresses scalability by adopting symbolic abstractions over physical skills expressed as governing physics equations with free parameters learnable from data. Such symbolic abstractions can be embedded into a planning domain for PDDL-style planning to arrive at multi-step plans. However, the hand-encoding of *exact* physics equations leads to brittleness during deployment in case of modelling errors.

In response, this study introduces a physics-informed skill acquisition and planning model that efficiently extends to multi-step reasoning tasks, offering advantages in (i) data efficiency, (ii) rapid physical reasoning, and (iii) reduced training time compared to methods relying on intricate physics simulations. Simulation time exhibits exponential growth when chaining skills, particularly in continuous state spaces. To address this, we leverage a limited number of physical demonstrations within the simulation engine to acquire *primitive* skills such as throwing, sliding, and rebounding from observed trajectories. Subsequently, we train a PINN-based predictive model. The acquired skills are then incorporated into a Monte Carlo Tree Search (MCTS) algorithm, facilitating the resolution of complex tasks through neural network simulations with minimal reliance on real-world simulations for fine-tuning (see Figure 1). Our modification of the MCTS algorithm (i) chooses to either enquire about the PINN-based coarse simulator or the fine simulator and (ii) accounts for the inherent sparsity of rewards commonly encountered in tasks involving physical reasoning. We call this approach PhyPlan. We designed simulated 3D physical reasoning tasks involving the Franka Emika arm using the Isaac Gym tool (5) to evaluate our approach. These tasks involve physical skills like sliding, throwing, and swinging. Each task requires the robot to get a box/ball/puck to reach the goal region by performing actions. We evaluate the agent in unseen scenarios based on how close it got the box/ball/puck to the goal. Empirical evaluation demonstrates that our proposed approach outperforms conventional methods, such as Deep Q-Networks (DQN), while requiring a notably reduced number of training simulations.

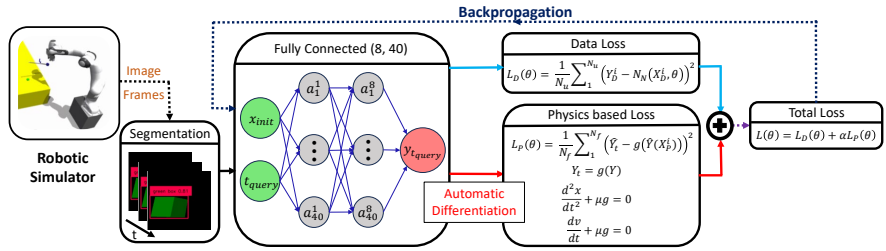
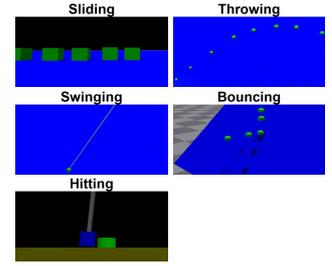

**Fig. 2. Physics Informed Skill Network Architecture.** A fully connected feed-forward network with eight hidden layers and 40 neurons per layer with ReLU activation. The network takes input initial conditions like position, velocity, etc. and the time the output value is desired. The loss function is composed of data loss and physics-based equation loss. Various skills are performed in the simulator, which feeds the data to the model either a) using image segmentation of successive frames (with perception) or b) directly from the simulator (without perception).

**Fig. 3. Dynamic Physical Skills.** Dynamic interactions by a manipulator we consider in this work. The agent's goal is to compose these skills to perform a high-level task.

## Related Work

Within robotics, several efforts have focused on learning physical skills using reinforcement learning or learning from demonstration. (6) use inverse RL to learn object insertion trajectories. (7) learn visuo-motor policies for throwing arbitrary objects. (8) focus on object transfer using catching. (9) learn to push strategies from observation. Instead of learning a predictive model for specific skills, others build on Dynamic Movement Primitives (10) as a generic skill representation to guide policy learning. Although effective for learning smooth (11) or periodic dynamics (12), the representation is less expressive for modelling contact or collision interactions. Although these approaches mentioned above demonstrate considerable success in learning short-horizon skills, their data requirements are high and scalability to long-horizon tasks is limited. Another body of work focuses on learning to compose skills for attaining goal-directed tasks. Closely related is the work of (4), where a planning framework is introduced for sequential manipulation using physical interactions such as hitting and throwing. Wherein this approach relies on exact physics equations, our approach learns a predictive model for skills using coarse physics and data, allowing more resilience to noise and accelerated training. Further, embedding skills into a model-based reinforcement learning approach facilitates solving multi-step tasks and adapting online to the environment's unmodeled aspects. The work of (7) introduces a trial and error method predicting residuals on top of control parameters specifying a skill parameterisation. (13) proposed to learn the latent reward function modelled as a Gaussian Process and trained using the rewards observed during training. The learnt function serves as a prior where corrections are learnt when the robot interacts in the real environment, an approach we adopt in this paper. A concurrent development bridging the realms of physics and data science is the Physics-Informed Neural Network (PINN) (14); the idea here is to impose regularisation on the latent space to ensure adherence to the governing physics. PINN offers distinct advantages, primarily driven by (i) its computational efficiency, thanks to the final model's neural network structure, and (ii) its remarkable generalisation capabilities derived from incorporating physics as a regularisation factor. Improvements over the vanilla PINN includes multi-fidelity PINN (15), gradient-enhanced PINN (16), and gradient-free PINN (17). In this work, we harness the power of PINN to model the foundational 'skills' necessary for executing complex physical reasoning 'tasks'.

## Problem Formulation

We consider a robot manipulator operating in a tabletop environment. Assume that the robot can position its end-effector at a desired configuration, grasp an object, move it while holding it, and release the object-oriented at a target pose. The environment consists of dynamic objects (box or a ball) and static objects such as a wedge, a flat surface (such as a table), a positioning plank (for instance, a bridge), a box-like object, a pendulum-like assembly capable of swinging a plan that rotates on a hinge. The dynamic object can interact with other objects, e.g., by sliding over a surface, colliding and rebounding from a wall, or more naturally under gravity (dropping vertically or falling as a projectile), finally coming to rest on a ground surface or inside a container-like object. Formally, let $s_t \in \mathcal{S}$ denote the world state at time $t$, observed as an image $i_t$ from a depth camera. Let $\mathcal{O}$ denote the set of objects in the environment, where each $o \in \mathcal{O}$ has a set of modes, $M(o)$ (degrees of freedom) for interacting with the environment. For example, a wedge may be po-sitioned at a coordinate $(x, y)$ and oriented as $q$, leading to $M(wedge) = \{x, y, q\}$. Similarly, the interaction modes for a pendulum can be the release angle $\theta$ and length $l$. For a projectile, it may be velocity $v$ and height of release $h$. We represent the action taken on state $s_t$ as $a_t \in \mathcal{A}$. Action outcomes are considered non-deterministic and modelled as a stochastic transition model $s_{t+1} \sim T(s_t, a_t)$, arising from execution in the real environment or a realistic physics simulation engine.

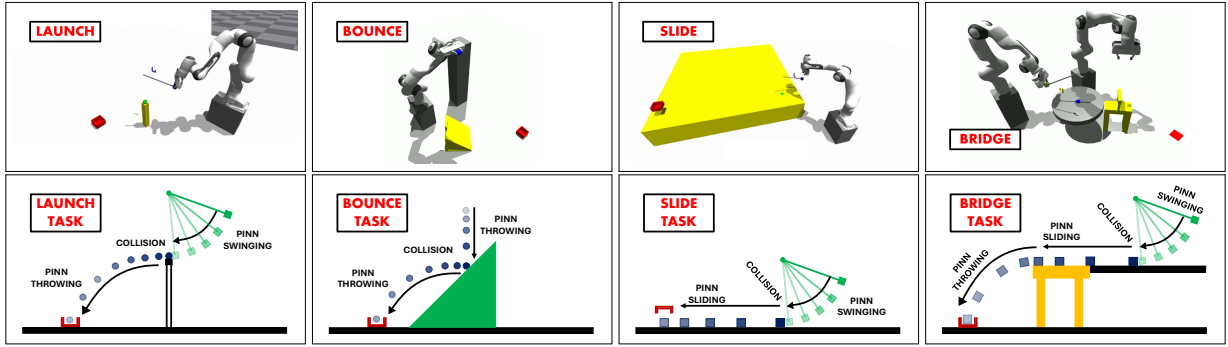

**Fig. 4. Benchmark Physical Reasoning Tasks** (named Launch, Bounce, Slide, and Bridge) in a 3D physics simulation environment with a simulated robot manipulator (above). The tasks are inspired by prior works in simplistic 2D environments in (1, 3). Typical strategies (below) for performing them require the robot to compose physical skills (Figure 3) to allow the moving object to reach a goal destination. See the labels indicating skills used in the figures below. For, to send a ball to a far-off-goal location (Bridge task), the robot can hit the ball with a pendulum, which causes sliding and later projectile motion to land in the goal region.

The robot's objective is to ensure that the dynamic object reaches a specified region of space (implicitly defined by a target container). We denote the goal region for the target object as $g$. Direct grasping and releasing the object may not be feasible, particularly when the goal region is far from the object. Hence, the robot may interact with objects as a tool, *i.e.*, position and release points in continuous space, such that when the object is released, the dynamic interactions facilitate reaching goal regions. For instance, a ball resting on a table may be hit by a pendulum, slide on a table, and fall into a target region via projectile motion. The goal regions are embedded in the reward model $R : \mathcal{S} \times \mathcal{A} \times \mathcal{S} \mapsto R$. This translates to a high reward when the object is contained within the goal region and low otherwise. In our setup, we obtain the reward $r_t$ after performing action $a_t$.

We learn a goal-conditioned policy $a_t \sim \pi_g(s_t)$, where $a_t$ is a tuple indicating the setting of interactive variables for each object, such as $a_t = \{M(o_1), M(o_2), \ldots, M(o_n)\}$. Directly learning this policy is challenging, mainly when the task requires sequential multi-step interaction over time, as this requires long-horizon temporal and causal reasoning as reaching goals with multiple tool-like objects spans a large space of possible trajectories. This work simplifies the long-horizon reasoning task by reducing the problem to myopic action optimisation. Here, we take action $a_t$ only based on $s_t$, obtain the reward, update our model, and reset decision-making. Expressly, we set the configuration or interactive variables of the dynamic objects to the desired values and release them. This is considered as a single "epoch". We do not consider future states like $s_{t+1}$. Hence, our setup is one of Contextual Multi-armed Bandits (C-MAB) (18). Our policy aims to minimise the regret $e = \frac{r - r*}{r}$, where $r$ is the maximum reward obtained by the agent in a simulation run, and $r*$ is the theoretical maximum reward. We observe that learning this policy requires simulation of actions. Such dynamic simulations are time-consuming and slow down the learning process. Hence, as we discuss next, we explore ways to accelerate learning using learnt physics simulators.

## Learning Physics-Informed Skill Models

Instead of learning a policy directly, we factor the learning problem above as one of learning a model that depicts how an object engages with the environment and then using this model to plan multi-step interactions to reach the intended goals. We postulate that a learnt physical interaction model would provide a strong prior for the planner and help them make informed decisions. We consider the problem of learning a model for physical skills such as *bouncing* a ball-like object off a wedge, *sliding* over an object, *swinging* a pendulum, *throwing* an object as a projectile and *hitting* an object with a pendulum. Here, we interpret a physical skill as a model that predicts the state trajectory of an object as it undergoes a dynamic interaction with another object.

Governing physics of such a skill can be represented as $Y_t = g(Y)$, where $Y$ represents the state variables, $Y_t$ represents the time derivative of the state variable, and $g(Y)$ represents a non-linear function of $Y$. For instance, skills like throwing can be modelled as projectile motion with a governing differential equation: $d^2y/dt^2 + g = 0$ representing motion in the vertical axis and $dx/dt - vx_{init} = 0$ representing motion in the horizontal plane. Here, $x$ and $y$ represent instantaneous position w.r.t. inertial frame in the horizontal and vertical direction, respectively, and $vx_{init}$ represents the initial horizontal velocity. Similarly, other skills like swinging can be modelled by pendulum dynamics and sliding as friction dynamics.

We introduce a neural network that predicts the object's state during dynamic interaction continuously parameterised by time. Such interactions can be simulated in a physics engine by using numerical integration schemes. However, since we aim to perform multi-step interactions, simulating outcomes during training is often intractable. Hence, we adopt a learning-based approach and learn a function $f(X, t) \to Y$ that predicts the object's state during dynamic interaction continuously parameterised by time. We represent $f(X, t)$ with a neural network and employ a physics-informed loss function to constrain the latent space, $L_P(\theta) = (1/N_f) \sum_1^{N_f} (\mathbb{R}(X_P^i, \theta))^2$, where $\mathbb{R}(X_P^i, \theta) := \hat{Y}_t - g(\hat{Y}(X_P^i))$ is residual of the governing physics defined before, and $X_P^i$ represents the $i^{th}$ collocation point. $\hat{Y}$ represents the neural network output, and $\hat{Y}_t(X_P^i)$ represents the derivative of the neural network predicted state variable obtained using automatic differentiation. For training the network in practice, we use both physics-based and data-driven loss functions, $L(\theta) = L_D(\theta) + L_P(\theta)$, where $L_D(\theta) = (1/N_u) \sum_1^{N_u} (Y_D^i - N_N(X_D^i, \theta))^2$. Figure 2 summarises the architecture of the Physics-Informed Skill Network. The physics-based loss enables the model to train from a sparse dataset ($N_u << N_f$), enabling rapid skills learning. In practice, the parameters associated with physics models are not known apriori. For example, the coefficient of friction is often not known. PINN treats the unknown parameter as additional trainable parameters (absorbed in $\theta$) and estimates the same by solving the aforementioned optimisation problem. In this work, PINN is employed to learn skills such as throwing, sliding, and swinging, and the trained PINN models are referred to as "Physics-Informed Skill Models". However, collision skills are learnt directly from data because of complex and intractable physics of the motion Figure 3 shows the skill sets considered in this paper.

## Physical Task Reasoning with Learnt Skills

The skill models developed so far allow the robot to predict how the object's trajectory will evolve during a dynamic physical interaction. This can be viewed as *single-step* reasoning. We now consider how to *compose* skills to solve a task requiring *multi-step* reasoning. For each task, we first figure out a chain of skills as described in Figure 4, and beginning with the initial state, we use a skill model to predict the state of the environment after each skill is executed. The state after the execution of the last skill is the final state of the environment, which is then evaluated for how near the ball/box/puck could get to the goal. The PINN-based Rollouts are semantic and approximately evaluate the model's performance based on how far the ball lands from the goal without involving the simulator.

We assume the chain of skills to perform the tasks, and PhyPlan finds the best action in each skill, such as the angle at which to swing the pendulum and the angle of the pendulum's plane in *Launch* task, the angle at which to place the wedge and the height of the ball in *Bounce* task, etc. We run an MCTS planning algorithm with `PINN-based Rollouts` (Algorithm 1) to search the continuous action space by uniformly sampling $\mathcal{D}$ (Discretization Factor) actions within the bounds of the space. The lines coloured blue in Algorithm 1 are the places where we differ from conventional MCTS. Also, $n_{action}$ and $v_{action}$ are the number of trials and the value of *action*, respectively. Algorithm 2 ex-

**Algorithm 1 : Physics-Informed MCTS**

**Input**: $node$   ▷ Each node corresponds to a state
**Output**: Reward
1: **if** $node$ is terminal **then**
2:   **return** $rv \leftarrow$ `PINN-Rollout`($node$)
3: **end if**
4: $\mathcal{A} \leftarrow$ actions considered in $node$
5: **if** $|\mathcal{A}| > \theta$ **then**
6:   $action \leftarrow \underset{a \in \mathcal{A}}{argmax}\, v_a + \alpha \sqrt{(\log \underset{a \in \mathcal{A}}{\sum} n_a)/n_a}$ ▷ Selection
7:   $newNode \leftarrow$ child of $node$ by taking $action$
8:   $rv \leftarrow$ MCTS($newNode$)
9:   $v_{action} \leftarrow (v_{action} \times n_{action} + rv)/(n_{action} + 1)$
10:   $n_{action} \leftarrow n_{action} + 1$
11: **else**
12:   $\mathcal{A} \leftarrow$ sample arms at $node$   ▷ Expansion
13:   **for** $a \in \mathcal{A}$ **do**
14:    $child \leftarrow$ child of node by taking arm $a$
15:    $v_a \leftarrow$ `PINN-Rollout`($child$)   ▷ Backpropogation
16:    $n_a \leftarrow 1$
17:   **end for**
18:   $rv \leftarrow$ `PINN-Rollout`($node$)   ▷ Simulation
19: **end if**
20: **return** $rv$

**Algorithm 2 : Adaptive Physical Task Reasoning**

**Input**: initial state, goal position, $num\_attempts$
**Output**: Regret
1: **for** attempt in $num\_attempts$ **do**
2:   Gaussian_Process.fit(`actions`, `rewards`) ▷ Adaptation
3:   $phy\_reward \leftarrow 0$
4:   **for** $\mathcal{K}$ iterations **do** ▷ Running Skill Model for $\mathcal{K}$ iterations
5:    $reward \leftarrow$ MCTS (root node)
6:    $action \leftarrow$ action sequence in Monte Carlo Tree for $reward$
7:    **if** $reward >$ phy_reward **then**
8:     phy_reward $\leftarrow reward$
9:     $best\_action \leftarrow action$
10:    **end if**
11:   **end for**
12:   `reward_sim` $\leftarrow$ execute `best_action` in simulation
13:   $best\_reward \leftarrow$ max($best\_reward$, $reward\_sim$)
14:   Append `best_action` to `actions`
15:   Append `reward_sim-phy_reward` to `rewards`
16: **end for**
17: **return** regret $\leftarrow \dfrac{(\text{opt\_reward} - \text{best\_reward})}{\text{opt\_reward}}$

pands the Monte Carlo Tree for a maximum number of iterations ($\mathcal{K}$) with a typical $\mathcal{D}$ and $\mathcal{K}$ at 20 and 10 respectively. Then, it executes the best action according to the skill model in the simulator. However, since the skill models could make systematic errors, we finetune our reward predictor by approximating the error as a Gaussian Process (GP). Given $t-1$ samples of action and error pairs, the GP provides a Gaussian posterior $\mathcal{N}(\mu_{t-1}, \sigma_{t-1})$ over the error of the next action $a_t$ for any action in the action space $\mathcal{A}$ (13). The MCTS uses this GP to correct the reward predictor in the search for the next action using the equation $rv_t = rv_t + \mu_{t-1}(a_t) + \beta^{1/2}\sigma_{t-1}(a_t)$, where $rv_t$ is the re-ward predicted by the skill model, $\mu_{t-1}(a_t)$ and $\sigma_{t-1}(a_t)$ are the mean and variance of the function value at $a$ at time $t-1$, and $\beta$ is a hyperparameter, we use $\beta = 0.25$.

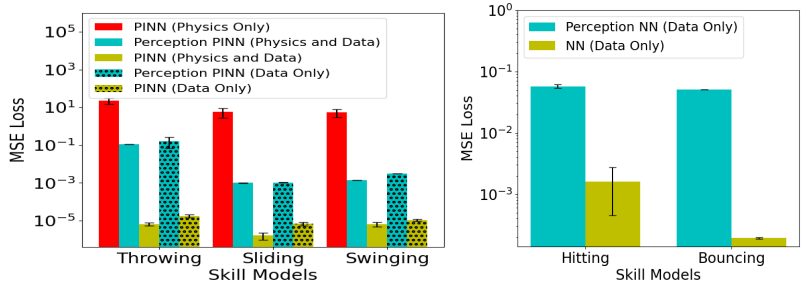
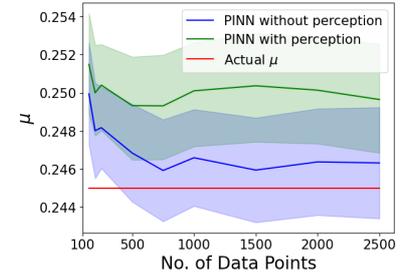

**Fig. 5. Skill Model Accuracy Comparison.** The figure highlights the prediction MSE loss on validation dataset. **Left:** PINN (Physics and Data) without perception performs better than other models highlighting advantage of incorporating guiding physics equation. Noise inherent in perception-based data can be attributed to less significant difference between PINN (Physics and Data) and PINN (Data only) with perception. **Right:** The performances of NN with and without perception are comparable within small bounds for skills trained via a data-driven approach.

**Fig. 6. Unknown Parameter:** Estimating $\mu$ in *sliding skill* with increasing no. of data points at 500 training cycles; PINN-based estimation achieves an error of $\sim 0.4\%$ without and $\sim 1\%$ with perception.

## Evaluation and Results

Through our evaluation, we first investigate the physics-informed skill model for the ability to predict object motions during dynamic interactions (such as sliding, hitting, throwing, etc.) and the data efficiency achieved compared to conventional physics-uninformed time series models. Then, we study the degree to which the proposed planner can learn to perform a physical reasoning task requiring a composition of learnt skills. Finally, we show, qualitatively, a simulated Franka Emika Robot performing the tasks.

### A. Experimentation Setup.

**Virtual Skill Learning:** We deploy Isaac Gym (5), a physics simulator and reinforcement learning framework for training robots and skill learning. The task involves learning physical skills like swinging, sliding, throwing, hitting and bouncing (Figure 3). Each task is simulated, and a Physics-Informed Skill model is trained over the data received. Consider the case of throwing a ball in the simulator, which provides velocity-displacement pair values of the projectile at different time instances as training input to the model. Similar experiments are done for each skill.

**Perception-based Skill Learning:** We also developed a perception-based pipeline to infer the object state directly from RGB-D data to replicate real lab-based settings. The experimentation scene in the simulator is captured via an inbuilt depth camera, which provides time series images of the experiment. On the gathered image dataset, segmentation techniques involving a combination of unsupervised detection methods like Segment Anything (19), Grounding Dino (20) and traditional methods like edge detection are used for object detection.

### B. Evaluation of Learnt Skills.
We evaluate the prediction accuracy of the proposed physics-informed skill network on simulation data and the physical parameter estimation in the unknown environment.

**Skill model accuracy** There are two types of proposed models based on the data collection regime employed: *Physics-Informed Neural Network* (Figure 2) a) *with perception* and b) *without perception*. For training the PINN model, actual data points and collocation or pseudo points (4 times the data points) are considered for optimising physics-based loss. The proposed models are compared against standard Feedforward network (NN) (same architecture as PINN but with only data-driven loss) (a) with perception and (b) without perception. Figure 5 indicates that the model with only physics loss performs poorly, and incorporating physics governing equations with the data loss allows stronger generalisation and improved data efficiency compared to general physics-uninformed time series models. PINN without perception is the most efficient skill-learning model. Although its performance degrades with perception-based training due to inherent noise in the data, PINN performs better than Perception NN.

**Unknown parameters and robustness** PINN-based skill models are particularly useful and efficient when the inherent variables of the environment are unknown, as depicted by Figure 6. PINN model is data efficient, which is highlighted by the fact that it requires very few data points (approximately 700-1000 points) to determine the actual $\mu$ value accurately. Note even though the perception-based approach is inherently noisy, there is a meagre difference between predicted $\mu$ values by PINN models with and without perception, depicting the robustness of our perception-based approach.

### C. Evaluation on Physical Reasoning Tasks.
**Dataset and Baselines** We created a benchmark data set composed of four tasks: Launch, Slide, Bounce, and Bridge, which require the robot to use physical skills sequentially to attain the goal. For example, the Bridge task requires a dynamic object like a ball to be positioned into a goal region. The task requires chaining multiple interactions with objects. The robot can interact with the environment multiple times, resetting the dynamic object after each complete rollout. Within each trial, the robot can set the objects in the desired configuration before releasing the dynamic object. The robot only observes an image of the scene. We also implemented the following baselines for comparisons: (i) *Random policy:* actions sampled

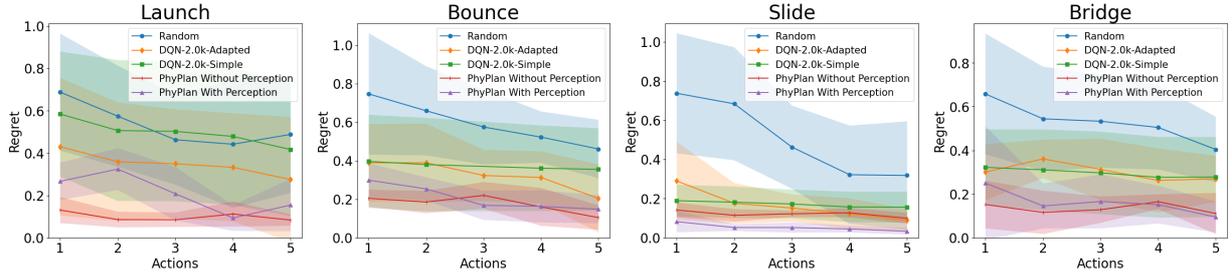

**Fig. 7. Task Solving Accuracy:** Comparing the performance of our approach and the baselines based on the minimum regret achieved vs the number of actions taken in the evaluation phase for each task. Lower is better. Our approach consistently beats the baselines.

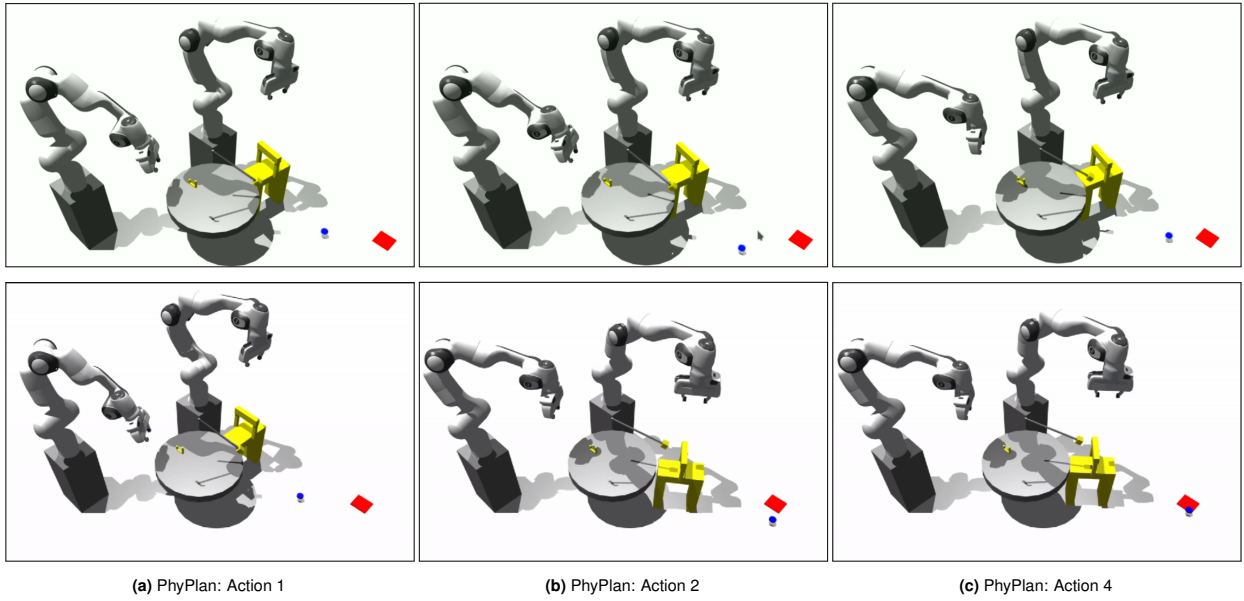

**(a)** PhyPlan: Action 1      **(b)** PhyPlan: Action 2      **(c)** PhyPlan: Action 4

**Fig. 8. Comparing PhyPlan with DQN-Adaptive.** Showing the actions performed by DQN-Adaptive vs those performed by PhyPlan in *Bridge* task. DQN does not use the bridge even after 11 actions unlike PhyPlan which uses it starting from the second action and perfectly aligns it in the fourth action to achieve the goal.

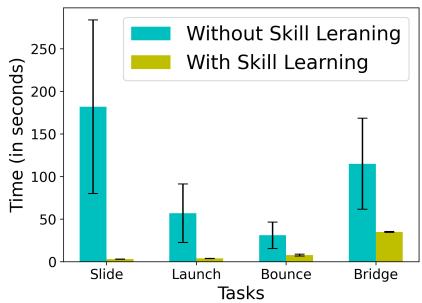

**Fig. 9.** PhyPlan's runtime is manifold less than that of planner without Physics-Informed skill models.

uniformly at random, (ii) *DQN-n-Simple:* a DQN baseline from (3) trained using random contexts (images) and random actions. The policy is extracted by selecting the highest scoring actions by sampling $10,000$ actions randomly and (iii) *DQN-n-Adapted*: similar to (ii) except for using a Gaussian Process to regress over rewards from prior rollouts.

**Task Learning Efficacy.** We built two versions of our model called *PhyPlan with perception* and *PhyPlan without perception*. The former uses perception-based PINN models and extracts the goal and ball positions from the task image. The latter uses PINN models without perception and obtains goal and ball positions directly from the simulator. Figure 7 illustrates the task learning efficiency measured as regret, defined in Section . The lower the regret, the better the performance. Both versions of PhyPlan achieve lower regret than the baselines. As the task complexity increases, the difference between PhyPlans' performance and the baselines widens. For instance, PhyPlan performs close to the baselines in the Sliding task, whereas it outperforms baselines by a high margin in the Bridge task. Further, *PhyPlan without perception* generally performs better than *PhyPlan with perception* because of the inherent noise in the Perception-based skill models and the errors in detecting the goal and ball positions. Figure 8 shows a qualitative comparison between PhyPlan and DQN-Adaptive in the *Bridge* task. PhyPlan demonstrates the capability of long-horizon reasoning in substantially less number of actions/trials. Further, Figure 9 compares our approach with its variant, which uses a full simulation engine during training instead of learnt skills. The results indicate the ben-efit of using the learnt model during planning. The proposed approach achieves significant runtime gains (reducing calls to the time-consuming physics engine) while achieving better task-solving ability.

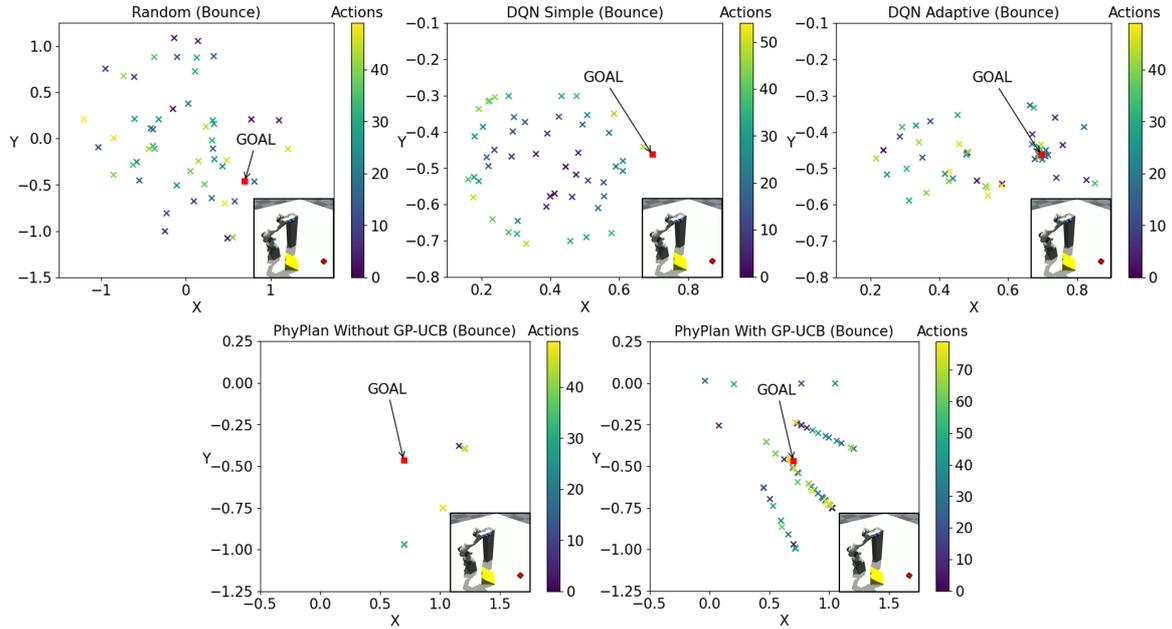

**Fig. 10. Visual Adaption of Reward Model:** Showing the coordinates of the goal and the ball for 50 iterations. Online Learning explores the action space exhaustively and converges to the optimal action by capturing the systematic errors made by the pre-trained models.

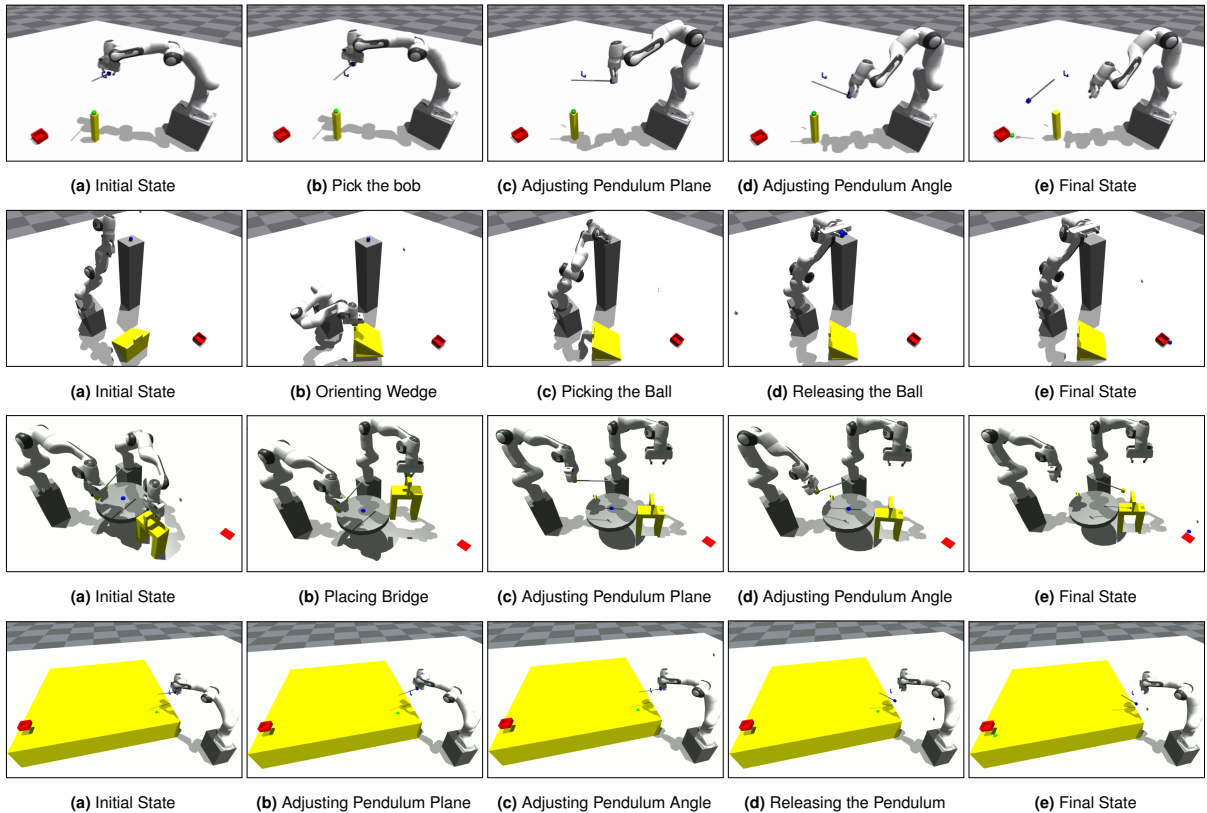

**Fig. 11. Simulated Robot Demonstrations.** Physical reasoning tasks learnt by a robot. See videos submitted in the supplementary material. **LAUNCH:** After 2 actions and 10 PINN-rollouts per action, the robot learns to correctly align the pendulum's plane and angle to throw the ball into the box. **BOUNCE:** After 3 actions and 10 PINN-rollouts per action, the robot learns to align the wedge correctly and releases the ball at an appropriate height to bounce it into the box. **BRIDGE:** After 5 actions and 20 PINN-rollouts per action, the robot learns to place the bridge in the path of the box and releases the pendulum from the correct location to slide the box to the goal. **SLIDE:** After 2 actions and 10 PINN-rollouts per action, the robot learns to correctly align the pendulum's plane and angle to slide the puck to the goal.

**Adaption of Reward Model.** We also evaluated the effectiveness of GP-UCB in adapting to the task execution environment, beginning with the pre-trained Physics-Informed Skill Models. Figure 10 illustrates the final positions where the dynamic object (the ball) lands relative to the goal after the robot performs the *Bounce* task. The Random model spans the whole action space, the *DQN-Simple* model is skewed as it does not learn online, the *DQN-Adaptive* model converges to the goal position in later iterations, *PhyPlan without GP-UCB* does not explore as it does not learn the prediction error, and *PhyPlan with GP-UCB* learns the prediction error to converge to the goal position quickly. Further, *PhyPlan with GP-UCB* appears to learn more systematically, exploring and refining within a confined action space.

Figure 11 demonstrates a simulated Franka Emika robot learning to perform the tasks. The robot interacts with the objects using crane grasping with a task space controller for trajectory tracking. The Figure shows the robot learning (i) the ball's release height and the wedge's pose in *Bounce* task; (ii) the pendulum's plane and angle in the *Launch* and the *Bridge* tasks; (iii) the bridge's pose in the *Bridge* task for the ball or puck to fall into the goal region. The policy is learned jointly for all objects, and the actions are delegated to the robot with the desired object in its kinematic range. Once the robots position the tools, the ball or the pendulum is dropped. Further, note that the robot learns to use the bridge effectively; a physical reasoning task reported earlier (1) to be challenging to learn for model-free methods, highlighting the PhyPlan's adaptability to long-horizon tasks. The results indicate the ability to generalise to new goal locations by using objects in the environments to aid task success.

## Conclusions

We consider the problem of training a robot to perform tasks requiring a compositional use of physical skills such as throwing, hitting, sliding, etc. Our approach leverages physics-governing equations to efficiently learn neural predictive models of physical skills from data. For goal-directed task reasoning, we embed learnt skills into an MCTS procedure that utilises the skill model to perform rapid rollouts during training (as opposed to using a full-scale physics simulation engine, which is slow in practice). Finally, we leverage the GP-UCB approach for online adaptations to compensate for modelling errors in skills by selective rollouts with real dynamics simulation. The approach is data efficient and performs better than the physics-uninformed DQN-based model. Future work will attempt to (a) learn the sequential composition of skills and plan the tasks *in-situ*, thus alleviating the assumption of learning all skills independently, (b) evaluate the model on a real robot, (c) investigate skill model for skills with complex or intractable physics, and (d) investigate PhyPlan's scalability to physical reasoning tasks involving large chain of physical skills.

# Appendix

This section includes the details of network architecture, perception methodology and the algorithm of Physics-informed rollouts with a visual description of the working of the algorithm. We also provide the links to access additional resources such as our code, dataset and website under the subsection *Code and Resources*.

**A. Skill Learning.**

In Table 1, we present a comprehensive overview of the Fully Connected Neural Network architecture implemented within various Physics-Informed Skill Networks. These networks integrate neural network models and physics-based governing equations specific to each skill. Additionally, Table 2 lists the inputs, outputs, and governing equations pertinent to each skill network. In this context, the subscript $init$ signifies the initial value of a physical parameter, and $t_{query}$ denotes the time instance for which the output quantity is required. The terms $v_{hor}$ and $v_{ver}$ are used to represent planar and vertical velocities, respectively, with the latter directed towards gravity (g). Key parameters such as the angle $\theta$ of a wedge in the bouncing skill and the masses $m_1$ and $m_2$ of colliding objects in the hitting skill are also inputs to their respective skill networks. Note that the skills like Bouncing and Hitting are exclusively trained through data-driven approaches, and hence, they do not incorporate any governing differential equations.

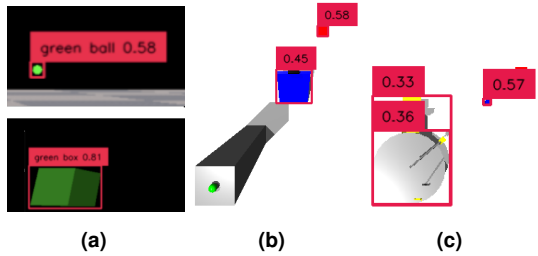

**Fig. 12.** Showing perception-based bounding box detection of a) ball being thrown and sliding box, to learn skills, b) wedge and goal in bounce task and c) goal in bridge task, for task inference

| Parameter | Value |
|---|---|
| Number of Layers | 10 (1 input, 8 hidden, 1 output) |
| Input Layer | Initial condition (pos, vel), $t_{query}$ |
| Output Layer | pos, vel at $t_{query}$ |
| Neurons in Hidden Layers | 40 |
| Activation Function (Hidden) | Tanh |
| Activation Function (Output) | None |
| Optimizer Type | L-BFGS |
| Learning Rate | 0.01 |
| Loss Function | MSE loss |
| Training Cycles | 6400 |
| Initialization | Xavier/Glorot Normal |

**Table 1.** General Neural Network Configuration used in different Physics-Informed Skill Networks

| Skill | Swinging | Sliding | Throwing | Bouncing | Hitting |
|---|---|---|---|---|---|
| Input | $(\theta_{init}, t_{query})$ | $(v_{init}, t_{query})$ | $(v_{hor_{init}}, v_{ver_{init}}, t_{query})$ | $(e, \theta, v_{ver_{init}}, v_{hor_{init}})$ | $(m_1, m_2, v_{init})$ |
| Output | $(\theta_{query}, \omega_{query})$ | $(x_{query}, v_{query})$ | $(v_{ver_{query}}, y_{query}, x_{query})$ | $(v_{ver_{query}}, v_{hor_{query}})$ | $v_{query}$ |
| ODE | $\frac{d^2\theta}{dt^2} + \frac{g*sin(\theta)}{l} = 0,$ $\frac{d\omega}{dt} + \frac{g*sin(\theta)}{l} = 0$ | $\frac{d^2x}{dt^2} + \mu g = 0,$ $\frac{dv}{dt} + \mu g = 0$ | $\frac{v_{ver_{init}}}{dt} + g = 0, \frac{d^2y}{d^2t} + g = 0,$ $\frac{dx}{dt} - v_{hor_{init}} = 0$ | None | None |

**Table 2. Physics Informed Skill Networks.** Showing the input, output, and ODE used to train different skill networks.

Figure 12 illustrates the object detection strategy employed in our Perception-based pipeline. Here, environmental scene images are processed using the 'Segment Anything' (19) and 'Grounding Dino' (20) models, which efficaciously detect targeted objects within a defined confidence. Figure 13 underscores the data efficiency of our employed Physics-Informed Neural Network in modeling physical skills. The PINN models exhibit lower Mean Squared Error (MSE) loss on the validation dataset with a relatively small number of training data points, compared to conventional Neural Network models.

**B. Algorithm.**

In Section 4 of the main paper, we delineate the intricacies of the algorithm employed in our study. Broadly, the *Adaptive Physical Task Reasoning* serves as the encompassing algorithm that invokes the *Physics-Informed MCTS* to acquire a nuanced reward model and adapt to the dynamic environment. The *Physics-Informed MCTS* performs semantic rollouts via the *PINN-Rollout* algorithm to approximate the value of actions, subsequently expanding the MCTS tree as per the insights gained. The procedural details of the semantic rollouts, featuring skill chaining with pre-trained Physics-Informed Skill Networks, are explained in Algorithm 3. Notably, the presumed knowledge of the sequence of physical skills in line 1 of the algorithm currently represents an operational assumption. We envisage mitigating this assumption in future iterations of our work, thereby enhancing the generalizability of this technique and reducing the need for manual intervention.

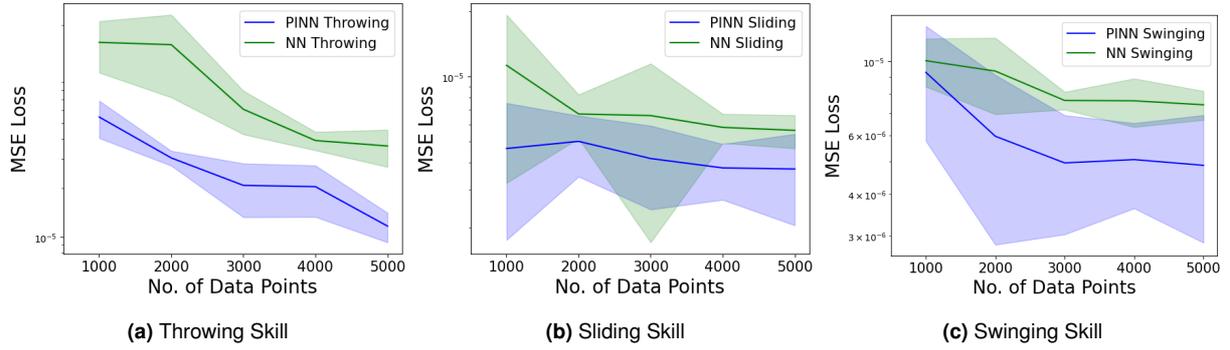

(a) Throwing Skill    (b) Sliding Skill    (c) Swinging Skill

**Fig. 13.** PINN displays superior data efficiency over NN, highlighted by lower MSE loss in various initial settings.

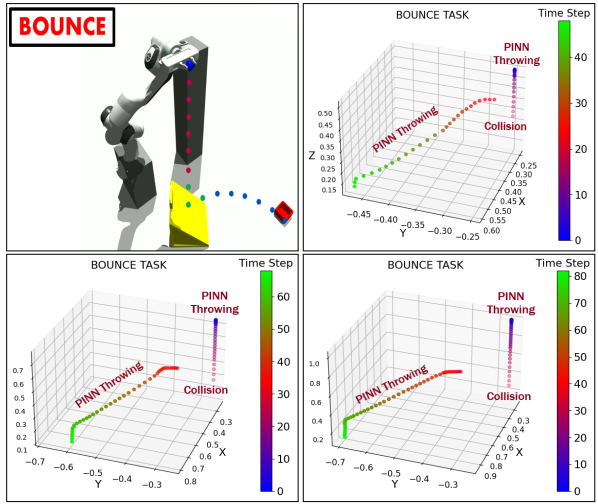

**Fig. 14.** Showing position of the ball predicted by a chain of skill networks at various time steps at a gap of 0.01 seconds ($t_{query}$) in *Bounce* Task. The agent varies the initial drop height (here $0.5, 0.75 \ and \ 1.0$) and expects the final resting position of the ball to evaluate the reward.

**Algorithm 3 : PINN-Rollout**

**Input**: $init\_node$    ▷ Each node corresponds to a state
**Output**: Reward
1: Skills ← Sequence of physical skills.
2: **for** s ∈ Actions **do**
3:    time_step ← $t_{query}$
4:    time ← time_step
5:    **while** $s \ is \ not \ finished$ **do**    ▷ Skill Chaining
6:      $new\_node$ ← Skill Model($init\_node$, time)
7:      time ← time + time_step
8:    **end while**
9:    $init\_node \leftarrow new\_node$    ▷ Transition between skills
10: **end for**
11: rv ← reward obtained
12: action ← sequence of actions chosen for rollout
13: $\mu, \sigma$ ← Gaussian_Process.predict(action)
14: **return** rv + $\left(\mu + \beta^{1/2} \times \sigma\right)$    ▷ Adding GP-UCB posterior

Figure 14 illustrates the Algorithm 3 in action. It shows the PINN-predicted trajectory of the ball for various initial heights tested by the agent. Notably, the ball's final resting position corresponds to an increasing trend with respect to the initial drop height. Specifically, the ball's terminal position for an initial height of $1.0$ exceeds that for $0.75$, which, in turn, surpasses the final position for an initial height of $0.5$. The PINN model demonstrates a notable capability to emulate the physical trajectory of the ball, aligning with the principles of physics governing the ball's motion.

### C. Code and Resources.

The code implementation associated with this research are available on GitHub at the following link: https://github.com/phyplan/PhyPlan
The datasets and additional resources are available on OneDrive at http://bit.ly/phyplan-resources
Please visit the our website at https://phyplan.github.io/ for further details.